\documentclass[sigconf]{acmart}

\AtBeginDocument{%
  \providecommand\BibTeX{{%
    \normalfont B\kern-0.5em{\scshape i\kern-0.25em b}\kern-0.8em\TeX}}}

\setcopyright{acmcopyright}
\copyrightyear{2018}
\acmYear{2018}
\acmDOI{10.1145/1122445.1122456}

\acmConference[Woodstock '18]{Woodstock '18: ACM Symposium on Neural
  Gaze Detection}{June 03--05, 2018}{Woodstock, NY}
\acmBooktitle{Woodstock '18: ACM Symposium on Neural Gaze Detection,
  June 03--05, 2018, Woodstock, NY}
\acmPrice{15.00}
\acmISBN{978-1-4503-XXXX-X/18/06}

\usepackage[utf8]{inputenc}
\usepackage[T1]{fontenc}

\usepackage{tabularx}
\usepackage{booktabs}
\usepackage{multirow}
\usepackage{makecell}
\usepackage{float}

\usepackage{graphicx}
\usepackage{caption}
\usepackage{listings}
\usepackage{algorithm,algorithmic}

\usepackage{xcolor}

\usepackage[nolist]{acronym}
\begin{acronym}[UML]
	\acro{AI}{Artificial Intelligence}
	\acro{BFS}{Breadth-First-Search}
	\acro{BoW}{Bag-of-Words}
	\acro{CSV}{Comma-Separated Values}
	\acro{CNN}{Convolutional Neural Network}
	\acro{DL}{Description logic}

	\acro{ER}{Entity Resolution}
	\acro{EM}{Expectation Maximization}
	\acro{EBMT}{Example-Based Machine Translation}
	\acro{EBNF}{Extended Backus--Naur Form}
	\acro{EL}{Entity Linking}
	\acro{FAO}{Food and Agriculture Organization of the United Nations}
	\acro{GIS}{Geographic Information Systems}
	\acro{GHO}{Global Health Observatory}
	\acro{GRU}{Gated recurrent unit}
	\acro{HDI}{Human Development Index}
	\acro{ICT}{Information and communication technologies}
	\acro{IFRS}{International Financial Reporting Standards}
	\acro{ICD}{International Classification of Diseases}
	\acro{IT}{Information Technology}
    \acro{KB}{Knowledge Base}
    \acro{KG}{Knowledge Graph}
    \acrodefplural{KG}{Knowledge Graphs}
    \acro{KGE}{Knowledge Graph Embeddings}
    \acro{KBSE}{Knowledge Base Semantic Embedding}
    \acro{KGC}{Knowledge Graph Completion}
	\acro{LR}  {Language Resource}
	\acro{LD}  {Linked Data}
	\acro{LLOD}  {Linguistic Linked Open Data}
	\acro{LIMES}{LInk discovery framework for MEtric Spaces}
	\acro{LS}  {Link Specifications}
	\acro{LDIF}{Linked Data Integration Framework}
	\acro{LGD} {LinkedGeoData}
	\acro{LOD} {Linked Open Data}
	\acro{LOV} {Linked Open Vocabularies}
	\acro{LSTM}{Long Short-Term Memories}
	\acro{MSE}{Mean Squared Error}
	\acro{ML}{Machine Learning}
	\acro{MR}{Machine Learning}
	\acro{MRR}{Mean Reciprocal Rank}
	
	\acro{NIF}{Natural Language Processing Interchange Format}
	\acro{NIF4OGGD}{NLP Interchange Format for Open German Governmental Data}
	\acro{NLP}{Natural Language Processing}
	\acro{NER}{Named Entity Recognition}
	\acro{NMT}{Neural Machine Translation}
	\acro{NN}{Neural Network}
	\acrodefplural{NN}{Neural Networks}
	\acro{NLG}{Natural Language Generation}
	\acro{NED}{Named Entity Disambiguation}
	\acro{NERD}{Named Entity Recognition and Disambiguation}
	\acro{NL}{Natural Language}
	\acro{NIF}{NLP Interchange Format}
	\acro{NIF4OGGD}{NLP Interchange Format for Open German Governmental Data}
	\acro{NLP}{Natural Language Processing}
	\acro{NER}{Named Entity Recognition}
	\acro{NEL}{Named Entity Linking}
	\acro{NE}{Named Entity}
	\acro{NN}{Neural Network}
	\acro{NLI}{Natural Language Inference}
	\acro{OSM}{OpenStreetMap}
	\acro{OWL}{Web Ontology Language}
	\acro{OOV}{out-of-vocabulary}
	\acro{PFM}{Pseudo-F-Measures}
	\acro{PSO}{Particle-Swarm Optimization}
	\acro{PBSMT}{Phrase-Based Statistical Machine Translation}
	
	\acro{QA}{Question Answering}
	\acro{RDF}{Resource Description Framework}
	\acro{RBMT}{Rule-Based Machine Translation}
	\acro{RNN}{Recurrent Neural Network}
	\acro{ReLU}{rectified linear unit}
	\acro{RDFS}{RDF Schema}
	\acro{SKOS}{Simple Knowledge Organization System}
	\acro{SPARQL}{SPARQL Protocol and RDF Query Language}
	\acro{SRL}{Statistical Relational Learning}
	\acro{SWT}{Semantic Web Technologies}
	\acro{SW}{Semantic Web}
	\acro{SMT}{Statistical Machine Translation}
	\acro{SWMT}{Semantic Web Machine Translation}
	\acro{SKOS}{Simple Knowledge Organization System}
	\acro{SPARQL}{SPARQL Protocol and RDF Query Language}
	\acro{SRL}{Statistical Relational Learning}
	\acro{SF}{surface forms}

    \acro{TBMT} {Transfer-Based Machine Translation}
	\acro{UML}{Unified Modeling Language}
	\acro{USL}{Ukrainian Sign Language}
	\acro{URI}{Uniform Resource Identifier}
	\acro{WHO}{World Health Organization}
	\acro{WKT}{Well-Known Text}
	\acro{W3C}{World Wide Web Consortium}
	\acro{WSD}{Word Sense Disambiguation}
	\acro{WMT}{Workshop on Machine Translation}
    \acro{XML}{Extensible Markup Language}
	\acro{YPLL}{Years of Potential Life Lost}

	\acro{AOS}{Agricultural Ontology Services}
	\acro{AGRIS}{Agricultural Science and Technology}
	\acro{API}{Application Programming Interface}
	\acro{AOS}{Agricultural Ontology Services}
	\acro{AGRIS}{Agricultural Science and Technology}
	\acro{API}{Application Programming Interface}
	\acro{A2KB}{Annotation to Knowledge Base}
	\acro{BPSO}{Binary Particle-Swarm Optimization}
	\acro{BPMLOD}{Best Practices for Multilingual Linked Open Data}
	\acro{BPSO}{Binary Particle-Swarm Optimization}
	\acro{BPMLOD}{Best Practices for Multilingual Linked Open Data}
	\acro{BFS}{Breadth-First-Search}
	\acro{BPE}{Byte Pair Encoding}
	\acro{BoW}{Bag-of-Words}
	\acro{CBD}{Concise Bounded Description}
	\acro{COG}{Content Oriented Guidelines}
	\acro{CSV}{Comma-Separated Values}
	\acro{CBMT}{Corpus-Based Machine Translation}
	\acro{CLIR}{Cross-Language Information Retrieval}
	\acro{DPSO}{Deterministic Particle-Swarm Optimization}
	\acro{DALY}{Disability Adjusted Life Year}

	\acro{ER}{Entity Resolution}
	\acro{EM}{Expectation Maximization}
	\acro{EBMT}{Example-Based Machine Translation}
	\acro{EBNF}{Extended Backus--Naur Form}
	\acro{EL}{Entity Linking}
	\acro{FAO}{Food and Agriculture Organization of the United Nations}
	\acro{GIS}{Geographic Information Systems}
	\acro{GHO}{Global Health Observatory}
	\acro{GRU}{Gated recurrent unit}
	\acro{HDI}{Human Development Index}
	\acro{ICT}{Information and communication technologies}
	\acro{IFRS}{International Financial Reporting Standards}
	\acro{ICD}{International Classification of Diseases}
	\acro{IT}{Information Technology}
	\acro{IRI}{International Resource Identifier}
    \acro{KB}{Knowledge Base}
    \acro{KG}{Knowledge Graph}
    \acro{KGE}{Knowledge Graph Embeddings}
    \acro{KBSE}{Knowledge Base Semantic Embedding}
	\acro{LR}  {Language Resource}
	\acro{LD}  {Linked Data}
	\acro{LLOD}  {Linguistic Linked Open Data}
	\acro{LIMES}{LInk discovery framework for MEtric Spaces}
	\acro{LS}  {Link Specifications}
	\acro{LDIF}{Linked Data Integration Framework}
	\acro{LGD} {LinkedGeoData}
	\acro{LOD} {Linked Open Data}
	\acro{LSTM}{Long Short-Term Memories}
	\acro{MSE}{Mean Squared Error}
	\acro{MWE}{Multiword Expressions}
	\acro{MT}{Machine Translation}
	\acro{ML}{Machine Learning}
	\acro{MR}{Machine Reading}
	\acro{MOS}{Manchester OWL Syntax}
	\acro{NIF}{Natural Language Processing Interchange Format}
	\acro{NIF4OGGD}{NLP Interchange Format for Open German Governmental Data}
	\acro{NLP}{Natural Language Processing}
	\acro{NER}{Named Entity Recognition}
	\acro{NMT}{Neural Machine Translation}
	\acro{NN}{Neural Network}
	\acro{NLG}{Natural Language Generation}
	\acro{NED}{Named Entity Disambiguation}
	\acro{NERD}{Named Entity Recognition and Disambiguation}
	\acro{NL}{Natural Language}
	\acro{NIF}{NLP Interchange Format}
	\acro{NIF4OGGD}{NLP Interchange Format for Open German Governmental Data}
	\acro{NLP}{Natural Language Processing}
	\acro{NER}{Named Entity Recognition}
	\acro{NEL}{Named Entity Linking}
	\acro{NE}{Named Entity}
	\acro{NN}{Neural Network}
	\acro{NLI}{Natural Language Inference}
	\acro{OSM}{OpenStreetMap}
	\acro{OWL}{Web Ontology Language}
	\acro{OOV}{out-of-vocabulary}
	\acro{PFM}{Pseudo-F-Measures}
	\acro{PSO}{Particle-Swarm Optimization}
	\acro{PBSMT}{Phrase-Based Statistical Machine Translation}
	
	\acro{QA}{Question Answering}
	\acro{RDF}{Resource Description Framework}
	\acro{RBMT}{Rule-Based Machine Translation}
	\acro{RNN}{Recurrent Neural Network}
	\acro{ReLU}{rectified linear unit}
	\acro{SKOS}{Simple Knowledge Organization System}
	\acro{SPARQL}{SPARQL Protocol and RDF Query Language}
	\acro{SRL}{Statistical Relational Learning}
	\acro{SWT}{Semantic Web Technologies}
	\acro{SW}{Semantic Web}
	\acro{SMT}{Statistical Machine Translation}
	\acro{SWMT}{Semantic Web Machine Translation}
	\acro{SKOS}{Simple Knowledge Organization System}
	\acro{SPARQL}{SPARQL Protocol and RDF Query Language}
	\acro{SRL}{Statistical Relational Learning}
	\acro{SF}{surface forms}
	\acro{SVM}{Support Vector Machines}

    \acro{TBMT} {Transfer-Based Machine Translation}
	\acro{UML}{Unified Modeling Language}
	\acro{USL}{Ukrainian Sign Language}
	\acro{URI}{Uniform Resource Identifier}
	\acro{WHO}{World Health Organization}
	\acro{WKT}{Well-Known Text}
	\acro{W3C}{World Wide Web Consortium}
	\acro{WSD}{Word Sense Disambiguation}
	\acro{WWW}{World Wide Web}
    \acro{XML}{Extensible Markup Language}
	\acro{YPLL}{Years of Potential Life Lost}

	\acro{AOS}{Agricultural Ontology Services}
	\acro{AGRIS}{Agricultural Science and Technology}
	\acro{API}{Application Programming Interface}
	\acro{BPSO}{Binary Particle-Swarm Optimization}
	\acro{BPMLOD}{Best Practices for Multilingual Linked Open Data}
	\acro{CBD}{Concise Bounded Description}
	\acro{COG}{Content Oriented Guidelines}
	\acro{CSV}{Comma-Separated Values}
	\acro{CBMT}{Corpus-Based Machine Translation}
	\acro{CLIR}{Cross-Language Information Retrieval}
	\acro{DPSO}{Deterministic Particle-Swarm Optimization}
	\acro{DALY}{Disability Adjusted Life Year}
	\acro{DRL}{Deep Reinforcement Learning}

	\acro{ER}{Entity Resolution}
	\acro{EM}{Expectation Maximization}
	\acro{EBMT}{Example-Based Machine Translation}
	\acro{EBNF}{Extended Backus--Naur Form}
	\acro{EL}{Entity Linking}
	\acro{FAO}{Food and Agriculture Organization of the United Nations}
	\acro{GIS}{Geographic Information Systems}
	\acro{GHO}{Global Health Observatory}
	\acro{HDI}{Human Development Index}
	\acro{ICT}{Information and communication technologies}
    \acro{KB}{Knowledge Base}
    \acro{KBSE}{Knowledge Base Semantic Embedding}
	\acro{LR}  {Language Resource}
	\acro{LD}  {Linked Data}
	\acro{LLOD}  {Linguistic Linked Open Data}
	\acro{LIMES}{LInk discovery framework for MEtric Spaces}
	\acro{LS}  {Link Specifications}
	\acro{LDIF}{Linked Data Integration Framework}
	\acro{LGD} {LinkedGeoData}
	\acro{LOD} {Linked Open Data}
	\acro{ML}{Machine Learning}
	\acro{MDP}{Markov Decision Process}
	\acro{MT}{Machine Translation}
	\acro{NIF}{Natural Language Processing Interchange Format}
	\acro{NIF4OGGD}{NLP Interchange Format for Open German Governmental Data}
	\acro{NLP}{Natural Language Processing}
	\acro{NER}{Named Entity Recognition}
	\acro{NMT}{Neural Machine Translation}
	\acro{NN}{Neural Network}
	\acro{NLG}{Natural Language Generation}
	\acro{NED}{Named Entity Disambiguation}
	\acro{NERD}{Named Entity Recognition and Disambiguation}
	\acro{NL}{Natural Language}
	\acro{OWL}{Web Ontology Language}
	\acro{QA}{Question Answering}
	\acro{RDF}{Resource Description Framework}
	\acro{RBMT}{Ru\-le-Ba\-sed Ma\-chi\-ne Trans\-la\-tion}
	\acro{REG}{Referring Expression Generation}
	\acro{SKOS}{Simple Knowledge Organization System}
	\acro{SPARQL}{SPARQL Protocol and RDF Query Language}
	\acro{SRL}{Statistical Relational Learning}
	\acro{SWT}{Semantic Web Technologies}
	\acro{SW}{Semantic Web}
	\acro{SMT}{Statistical Machine Translation}
	\acro{SWMT}{Semantic Web Machine Translation}

    \acro{TBMT} {Transfer-Based Machine Translation}
	\acro{VS}{Version Space}

	\acro{WHO}{World Health Organization}
	\acro{WKT}{Well-Known Text}
	\acro{W3C}{World Wide Web Consortium}
	\acro{WSD}{Word Sense Disambiguation}
    \acro{XML}{Extensible Markup Language}
	\acro{YPLL}{Years of Potential Life Lost}

	\acro{AOS}{Agricultural Ontology Services}
	\acro{AGRIS}{Agricultural Science and Technology}
	\acro{API}{Application Programming Interface}
	\acro{AOS}{Agricultural Ontology Services}
	\acro{AGRIS}{Agricultural Science and Technology}
	\acro{API}{Application Programming Interface}
	\acro{A2KB}{Annotation to Knowledge Base}
	\acro{BPSO}{Binary Particle-Swarm Optimization}
	\acro{BPMLOD}{Best Practices for Multilingual Linked Open Data}
	\acro{BPSO}{Binary Particle-Swarm Optimization}
	\acro{BPMLOD}{Best Practices for Multilingual Linked Open Data}
	\acro{BFS}{Breadth-First-Search}
	\acro{BPE}{Byte Pair Encoding}
	\acro{BoW}{Bag-of-Words}
	\acro{CBD}{Concise Bounded Description}
	\acro{COG}{Content Oriented Guidelines}
	\acro{CSV}{Comma-Separated Values}
	\acro{CBMT}{Corpus-Based Machine Translation}
	\acro{CLIR}{Cross-Language Information Retrieval}
	\acro{DPSO}{Deterministic Particle-Swarm Optimization}
	\acro{DALY}{Disability Adjusted Life Year}

	\acro{ER}{Entity Resolution}
	\acro{EM}{Expectation Maximization}
	\acro{EBMT}{Example-Based Machine Translation}
	\acro{EBNF}{Extended Backus--Naur Form}
	\acro{EL}{Entity Linking}
	\acro{FAO}{Food and Agriculture Organization of the United Nations}
	\acro{GIS}{Geographic Information Systems}
	\acro{GHO}{Global Health Observatory}
	\acro{GRU}{Gated recurrent unit}
	\acro{HDI}{Human Development Index}
	\acro{ICT}{Information and communication technologies}
	\acro{IFRS}{International Financial Reporting Standards}
	\acro{ICD}{International Classification of Diseases}
	\acro{IT}{Information Technology}
    \acro{KB}{Knowledge Base}
    \acro{KG}{Knowledge Graph}
    \acro{KGE}{Knowledge Graph Embedding}
	\acro{LR}  {Language Resource}
	\acro{LD}  {Linked Data}
	\acro{LLOD}  {Linguistic Linked Open Data}
	\acro{LIMES}{LInk discovery framework for MEtric Spaces}
	\acro{LS}  {Link Specifications}
	\acro{LDIF}{Linked Data Integration Framework}
	\acro{LGD} {LinkedGeoData}
	\acro{LOD} {Linked Open Data}
	\acro{LSTM}{Long Short-Term Memories}
	\acro{MSE}{Mean Squared Error}
	\acro{MWE}{Multiword Expressions}
	\acro{MT}{Machine Translation}
	\acro{ML}{Machine Learning}
	\acro{MR}{Machine Reading}
	\acro{NIF}{Natural Language Processing Interchange Format}
	\acro{NIF4OGGD}{NLP Interchange Format for Open German Governmental Data}
	\acro{NLP}{Natural Language Processing}
	\acro{NER}{Named Entity Recognition}
	\acro{NMT}{Neural Machine Translation}
	\acro{NN}{Neural Network}
	\acro{NLG}{Natural Language Generation}
	\acro{NED}{Named Entity Disambiguation}
	\acro{NERD}{Named Entity Recognition and Disambiguation}
	\acro{NL}{Natural Language}
	\acro{NIF}{NLP Interchange Format}
	\acro{NIF4OGGD}{NLP Interchange Format for Open German Governmental Data}
	\acro{NLP}{Natural Language Processing}
	\acro{NER}{Named Entity Recognition}
	\acro{NEL}{Named Entity Linking}
	\acro{NE}{Named Entity}
	\acro{NN}{Neural Network}
	\acro{NLI}{Natural Language Inference}
	\acro{OSM}{OpenStreetMap}
	\acro{OWL}{Web Ontology Language}
	\acro{OOV}{out-of-vocabulary}
	\acro{PFM}{Pseudo-F-Measures}
	\acro{PSO}{Particle-Swarm Optimization}
	\acro{PBSMT}{Phrase-Based Statistical Machine Translation}
	
	\acro{QA}{Question Answering}
	\acro{RL}{Reinforcement Learning}
	\acro{RDF}{Resource Description Framework}
	\acro{RNN}{Recurrent Neural Network}
	\acro{ReLU}{rectified linear unit}
	
	\acro{SKOS}{Simple Knowledge Organization System}
	\acro{SWT}{Semantic Web Technologies}
	\acro{SW}{Semantic Web}
	\acro{SPARQL}{SPARQL Protocol and RDF Query Language}
	\acro{SRL}{Statistical Relational Learning}
	\acro{SVM}{Support Vector Machines}
    \acro{SGD}{Stochastic Gradient Descent}

    \acro{TBMT} {Transfer-Based Machine Translation}
	\acro{UML}{Unified Modeling Language}
	\acro{USL}{Ukrainian Sign Language}
	\acro{URI}{Uniform Resource Identifier}
	\acro{WHO}{World Health Organization}
	\acro{WKT}{Well-Known Text}
	\acro{W3C}{World Wide Web Consortium}
	\acro{WSD}{Word Sense Disambiguation}
    \acro{XML}{Extensible Markup Language}
	\acro{YPLL}{Years of Potential Life Lost}
\end{acronym}  

\usepackage{esvect}
\usepackage{xspace}
\usepackage{mathrsfs}
\usepackage{enumerate}
\usepackage{nicefrac}
\usepackage{lscape}

\usepackage{cleveref}

\usepackage[nolist]{acronym}
\begin{acronym}[UML]
	\acro{AI}{Artificial Intelligence}
	\acro{BFS}{Breadth-First-Search}
	\acro{BoW}{Bag-of-Words}
	\acro{CSV}{Comma-Separated Values}
	\acro{CNN}{Convolutional Neural Network}
	\acro{DL}{Description logic}

	\acro{ER}{Entity Resolution}
	\acro{EM}{Expectation Maximization}
	\acro{EBMT}{Example-Based Machine Translation}
	\acro{EBNF}{Extended Backus--Naur Form}
	\acro{EL}{Entity Linking}
	\acro{FAO}{Food and Agriculture Organization of the United Nations}
	\acro{GIS}{Geographic Information Systems}
	\acro{GHO}{Global Health Observatory}
	\acro{GRU}{Gated recurrent unit}
	\acro{HDI}{Human Development Index}
	\acro{ICT}{Information and communication technologies}
	\acro{IFRS}{International Financial Reporting Standards}
	\acro{ICD}{International Classification of Diseases}
	\acro{IT}{Information Technology}
    \acro{KB}{Knowledge Base}
    \acro{KG}{Knowledge Graph}
    \acrodefplural{KG}{Knowledge Graphs}
    \acro{KGE}{Knowledge Graph Embeddings}
    \acro{KBSE}{Knowledge Base Semantic Embedding}
    \acro{KGC}{Knowledge Graph Completion}
	\acro{LR}  {Language Resource}
	\acro{LD}  {Linked Data}
	\acro{LLOD}  {Linguistic Linked Open Data}
	\acro{LIMES}{LInk discovery framework for MEtric Spaces}
	\acro{LS}  {Link Specifications}
	\acro{LDIF}{Linked Data Integration Framework}
	\acro{LGD} {LinkedGeoData}
	\acro{LOD} {Linked Open Data}
	\acro{LOV} {Linked Open Vocabularies}
	\acro{LSTM}{Long Short-Term Memories}
	\acro{MSE}{Mean Squared Error}
	\acro{ML}{Machine Learning}
	\acro{MR}{Machine Learning}
	\acro{MRR}{Mean Reciprocal Rank}
	
	\acro{NIF}{Natural Language Processing Interchange Format}
	\acro{NIF4OGGD}{NLP Interchange Format for Open German Governmental Data}
	\acro{NLP}{Natural Language Processing}
	\acro{NER}{Named Entity Recognition}
	\acro{NMT}{Neural Machine Translation}
	\acro{NN}{Neural Network}
	\acrodefplural{NN}{Neural Networks}
	\acro{NLG}{Natural Language Generation}
	\acro{NED}{Named Entity Disambiguation}
	\acro{NERD}{Named Entity Recognition and Disambiguation}
	\acro{NL}{Natural Language}
	\acro{NIF}{NLP Interchange Format}
	\acro{NIF4OGGD}{NLP Interchange Format for Open German Governmental Data}
	\acro{NLP}{Natural Language Processing}
	\acro{NER}{Named Entity Recognition}
	\acro{NEL}{Named Entity Linking}
	\acro{NE}{Named Entity}
	\acro{NN}{Neural Network}
	\acro{NLI}{Natural Language Inference}
	\acro{OSM}{OpenStreetMap}
	\acro{OWL}{Web Ontology Language}
	\acro{OOV}{out-of-vocabulary}
	\acro{PFM}{Pseudo-F-Measures}
	\acro{PSO}{Particle-Swarm Optimization}
	\acro{PBSMT}{Phrase-Based Statistical Machine Translation}
	
	\acro{QA}{Question Answering}
	\acro{RDF}{Resource Description Framework}
	\acro{RBMT}{Rule-Based Machine Translation}
	\acro{RNN}{Recurrent Neural Network}
	\acro{ReLU}{rectified linear unit}
	\acro{RDFS}{RDF Schema}
	\acro{SKOS}{Simple Knowledge Organization System}
	\acro{SPARQL}{SPARQL Protocol and RDF Query Language}
	\acro{SRL}{Statistical Relational Learning}
	\acro{SWT}{Semantic Web Technologies}
	\acro{SW}{Semantic Web}
	\acro{SMT}{Statistical Machine Translation}
	\acro{SWMT}{Semantic Web Machine Translation}
	\acro{SKOS}{Simple Knowledge Organization System}
	\acro{SPARQL}{SPARQL Protocol and RDF Query Language}
	\acro{SRL}{Statistical Relational Learning}
	\acro{SF}{surface forms}

    \acro{TBMT} {Transfer-Based Machine Translation}
	\acro{UML}{Unified Modeling Language}
	\acro{USL}{Ukrainian Sign Language}
	\acro{URI}{Uniform Resource Identifier}
	\acro{WHO}{World Health Organization}
	\acro{WKT}{Well-Known Text}
	\acro{W3C}{World Wide Web Consortium}
	\acro{WSD}{Word Sense Disambiguation}
	\acro{WMT}{Workshop on Machine Translation}
    \acro{XML}{Extensible Markup Language}
	\acro{YPLL}{Years of Potential Life Lost}

	\acro{AOS}{Agricultural Ontology Services}
	\acro{AGRIS}{Agricultural Science and Technology}
	\acro{API}{Application Programming Interface}
	\acro{AOS}{Agricultural Ontology Services}
	\acro{AGRIS}{Agricultural Science and Technology}
	\acro{API}{Application Programming Interface}
	\acro{A2KB}{Annotation to Knowledge Base}
	\acro{BPSO}{Binary Particle-Swarm Optimization}
	\acro{BPMLOD}{Best Practices for Multilingual Linked Open Data}
	\acro{BPSO}{Binary Particle-Swarm Optimization}
	\acro{BPMLOD}{Best Practices for Multilingual Linked Open Data}
	\acro{BFS}{Breadth-First-Search}
	\acro{BPE}{Byte Pair Encoding}
	\acro{BoW}{Bag-of-Words}
	\acro{CBD}{Concise Bounded Description}
	\acro{COG}{Content Oriented Guidelines}
	\acro{CSV}{Comma-Separated Values}
	\acro{CBMT}{Corpus-Based Machine Translation}
	\acro{CLIR}{Cross-Language Information Retrieval}
	\acro{DPSO}{Deterministic Particle-Swarm Optimization}
	\acro{DALY}{Disability Adjusted Life Year}

	\acro{ER}{Entity Resolution}
	\acro{EM}{Expectation Maximization}
	\acro{EBMT}{Example-Based Machine Translation}
	\acro{EBNF}{Extended Backus--Naur Form}
	\acro{EL}{Entity Linking}
	\acro{FAO}{Food and Agriculture Organization of the United Nations}
	\acro{GIS}{Geographic Information Systems}
	\acro{GHO}{Global Health Observatory}
	\acro{GRU}{Gated recurrent unit}
	\acro{HDI}{Human Development Index}
	\acro{ICT}{Information and communication technologies}
	\acro{IFRS}{International Financial Reporting Standards}
	\acro{ICD}{International Classification of Diseases}
	\acro{IT}{Information Technology}
	\acro{IRI}{International Resource Identifier}
    \acro{KB}{Knowledge Base}
    \acro{KG}{Knowledge Graph}
    \acro{KGE}{Knowledge Graph Embeddings}
    \acro{KBSE}{Knowledge Base Semantic Embedding}
	\acro{LR}  {Language Resource}
	\acro{LD}  {Linked Data}
	\acro{LLOD}  {Linguistic Linked Open Data}
	\acro{LIMES}{LInk discovery framework for MEtric Spaces}
	\acro{LS}  {Link Specifications}
	\acro{LDIF}{Linked Data Integration Framework}
	\acro{LGD} {LinkedGeoData}
	\acro{LOD} {Linked Open Data}
	\acro{LSTM}{Long Short-Term Memories}
	\acro{MSE}{Mean Squared Error}
	\acro{MWE}{Multiword Expressions}
	\acro{MT}{Machine Translation}
	\acro{ML}{Machine Learning}
	\acro{MR}{Machine Reading}
	\acro{MOS}{Manchester OWL Syntax}
	\acro{NIF}{Natural Language Processing Interchange Format}
	\acro{NIF4OGGD}{NLP Interchange Format for Open German Governmental Data}
	\acro{NLP}{Natural Language Processing}
	\acro{NER}{Named Entity Recognition}
	\acro{NMT}{Neural Machine Translation}
	\acro{NN}{Neural Network}
	\acro{NLG}{Natural Language Generation}
	\acro{NED}{Named Entity Disambiguation}
	\acro{NERD}{Named Entity Recognition and Disambiguation}
	\acro{NL}{Natural Language}
	\acro{NIF}{NLP Interchange Format}
	\acro{NIF4OGGD}{NLP Interchange Format for Open German Governmental Data}
	\acro{NLP}{Natural Language Processing}
	\acro{NER}{Named Entity Recognition}
	\acro{NEL}{Named Entity Linking}
	\acro{NE}{Named Entity}
	\acro{NN}{Neural Network}
	\acro{NLI}{Natural Language Inference}
	\acro{OSM}{OpenStreetMap}
	\acro{OWL}{Web Ontology Language}
	\acro{OOV}{out-of-vocabulary}
	\acro{PFM}{Pseudo-F-Measures}
	\acro{PSO}{Particle-Swarm Optimization}
	\acro{PBSMT}{Phrase-Based Statistical Machine Translation}
	
	\acro{QA}{Question Answering}
	\acro{RDF}{Resource Description Framework}
	\acro{RBMT}{Rule-Based Machine Translation}
	\acro{RNN}{Recurrent Neural Network}
	\acro{ReLU}{rectified linear unit}
	\acro{SKOS}{Simple Knowledge Organization System}
	\acro{SPARQL}{SPARQL Protocol and RDF Query Language}
	\acro{SRL}{Statistical Relational Learning}
	\acro{SWT}{Semantic Web Technologies}
	\acro{SW}{Semantic Web}
	\acro{SMT}{Statistical Machine Translation}
	\acro{SWMT}{Semantic Web Machine Translation}
	\acro{SKOS}{Simple Knowledge Organization System}
	\acro{SPARQL}{SPARQL Protocol and RDF Query Language}
	\acro{SRL}{Statistical Relational Learning}
	\acro{SF}{surface forms}
	\acro{SVM}{Support Vector Machines}

    \acro{TBMT} {Transfer-Based Machine Translation}
	\acro{UML}{Unified Modeling Language}
	\acro{USL}{Ukrainian Sign Language}
	\acro{URI}{Uniform Resource Identifier}
	\acro{WHO}{World Health Organization}
	\acro{WKT}{Well-Known Text}
	\acro{W3C}{World Wide Web Consortium}
	\acro{WSD}{Word Sense Disambiguation}
	\acro{WWW}{World Wide Web}
    \acro{XML}{Extensible Markup Language}
	\acro{YPLL}{Years of Potential Life Lost}

	\acro{AOS}{Agricultural Ontology Services}
	\acro{AGRIS}{Agricultural Science and Technology}
	\acro{API}{Application Programming Interface}
	\acro{BPSO}{Binary Particle-Swarm Optimization}
	\acro{BPMLOD}{Best Practices for Multilingual Linked Open Data}
	\acro{CBD}{Concise Bounded Description}
	\acro{COG}{Content Oriented Guidelines}
	\acro{CSV}{Comma-Separated Values}
	\acro{CBMT}{Corpus-Based Machine Translation}
	\acro{CLIR}{Cross-Language Information Retrieval}
	\acro{DPSO}{Deterministic Particle-Swarm Optimization}
	\acro{DALY}{Disability Adjusted Life Year}
	\acro{DRL}{Deep Reinforcement Learning}

	\acro{ER}{Entity Resolution}
	\acro{EM}{Expectation Maximization}
	\acro{EBMT}{Example-Based Machine Translation}
	\acro{EBNF}{Extended Backus--Naur Form}
	\acro{EL}{Entity Linking}
	\acro{FAO}{Food and Agriculture Organization of the United Nations}
	\acro{GIS}{Geographic Information Systems}
	\acro{GHO}{Global Health Observatory}
	\acro{HDI}{Human Development Index}
	\acro{ICT}{Information and communication technologies}
    \acro{KB}{Knowledge Base}
    \acro{KBSE}{Knowledge Base Semantic Embedding}
	\acro{LR}  {Language Resource}
	\acro{LD}  {Linked Data}
	\acro{LLOD}  {Linguistic Linked Open Data}
	\acro{LIMES}{LInk discovery framework for MEtric Spaces}
	\acro{LS}  {Link Specifications}
	\acro{LDIF}{Linked Data Integration Framework}
	\acro{LGD} {LinkedGeoData}
	\acro{LOD} {Linked Open Data}
	\acro{ML}{Machine Learning}
	\acro{MDP}{Markov Decision Process}
	\acro{MT}{Machine Translation}
	\acro{NIF}{Natural Language Processing Interchange Format}
	\acro{NIF4OGGD}{NLP Interchange Format for Open German Governmental Data}
	\acro{NLP}{Natural Language Processing}
	\acro{NER}{Named Entity Recognition}
	\acro{NMT}{Neural Machine Translation}
	\acro{NN}{Neural Network}
	\acro{NLG}{Natural Language Generation}
	\acro{NED}{Named Entity Disambiguation}
	\acro{NERD}{Named Entity Recognition and Disambiguation}
	\acro{NL}{Natural Language}
	\acro{OWL}{Web Ontology Language}
	\acro{QA}{Question Answering}
	\acro{RDF}{Resource Description Framework}
	\acro{RBMT}{Ru\-le-Ba\-sed Ma\-chi\-ne Trans\-la\-tion}
	\acro{REG}{Referring Expression Generation}
	\acro{SKOS}{Simple Knowledge Organization System}
	\acro{SPARQL}{SPARQL Protocol and RDF Query Language}
	\acro{SRL}{Statistical Relational Learning}
	\acro{SWT}{Semantic Web Technologies}
	\acro{SW}{Semantic Web}
	\acro{SMT}{Statistical Machine Translation}
	\acro{SWMT}{Semantic Web Machine Translation}

    \acro{TBMT} {Transfer-Based Machine Translation}
	\acro{VS}{Version Space}

	\acro{WHO}{World Health Organization}
	\acro{WKT}{Well-Known Text}
	\acro{W3C}{World Wide Web Consortium}
	\acro{WSD}{Word Sense Disambiguation}
    \acro{XML}{Extensible Markup Language}
	\acro{YPLL}{Years of Potential Life Lost}

	\acro{AOS}{Agricultural Ontology Services}
	\acro{AGRIS}{Agricultural Science and Technology}
	\acro{API}{Application Programming Interface}
	\acro{AOS}{Agricultural Ontology Services}
	\acro{AGRIS}{Agricultural Science and Technology}
	\acro{API}{Application Programming Interface}
	\acro{A2KB}{Annotation to Knowledge Base}
	\acro{BPSO}{Binary Particle-Swarm Optimization}
	\acro{BPMLOD}{Best Practices for Multilingual Linked Open Data}
	\acro{BPSO}{Binary Particle-Swarm Optimization}
	\acro{BPMLOD}{Best Practices for Multilingual Linked Open Data}
	\acro{BFS}{Breadth-First-Search}
	\acro{BPE}{Byte Pair Encoding}
	\acro{BoW}{Bag-of-Words}
	\acro{CBD}{Concise Bounded Description}
	\acro{COG}{Content Oriented Guidelines}
	\acro{CSV}{Comma-Separated Values}
	\acro{CBMT}{Corpus-Based Machine Translation}
	\acro{CLIR}{Cross-Language Information Retrieval}
	\acro{DPSO}{Deterministic Particle-Swarm Optimization}
	\acro{DALY}{Disability Adjusted Life Year}

	\acro{ER}{Entity Resolution}
	\acro{EM}{Expectation Maximization}
	\acro{EBMT}{Example-Based Machine Translation}
	\acro{EBNF}{Extended Backus--Naur Form}
	\acro{EL}{Entity Linking}
	\acro{FAO}{Food and Agriculture Organization of the United Nations}
	\acro{GIS}{Geographic Information Systems}
	\acro{GHO}{Global Health Observatory}
	\acro{GRU}{Gated recurrent unit}
	\acro{HDI}{Human Development Index}
	\acro{ICT}{Information and communication technologies}
	\acro{IFRS}{International Financial Reporting Standards}
	\acro{ICD}{International Classification of Diseases}
	\acro{IT}{Information Technology}
    \acro{KB}{Knowledge Base}
    \acro{KG}{Knowledge Graph}
    \acro{KGE}{Knowledge Graph Embedding}
	\acro{LR}  {Language Resource}
	\acro{LD}  {Linked Data}
	\acro{LLOD}  {Linguistic Linked Open Data}
	\acro{LIMES}{LInk discovery framework for MEtric Spaces}
	\acro{LS}  {Link Specifications}
	\acro{LDIF}{Linked Data Integration Framework}
	\acro{LGD} {LinkedGeoData}
	\acro{LOD} {Linked Open Data}
	\acro{LSTM}{Long Short-Term Memories}
	\acro{MSE}{Mean Squared Error}
	\acro{MWE}{Multiword Expressions}
	\acro{MT}{Machine Translation}
	\acro{ML}{Machine Learning}
	\acro{MR}{Machine Reading}
	\acro{NIF}{Natural Language Processing Interchange Format}
	\acro{NIF4OGGD}{NLP Interchange Format for Open German Governmental Data}
	\acro{NLP}{Natural Language Processing}
	\acro{NER}{Named Entity Recognition}
	\acro{NMT}{Neural Machine Translation}
	\acro{NN}{Neural Network}
	\acro{NLG}{Natural Language Generation}
	\acro{NED}{Named Entity Disambiguation}
	\acro{NERD}{Named Entity Recognition and Disambiguation}
	\acro{NL}{Natural Language}
	\acro{NIF}{NLP Interchange Format}
	\acro{NIF4OGGD}{NLP Interchange Format for Open German Governmental Data}
	\acro{NLP}{Natural Language Processing}
	\acro{NER}{Named Entity Recognition}
	\acro{NEL}{Named Entity Linking}
	\acro{NE}{Named Entity}
	\acro{NN}{Neural Network}
	\acro{NLI}{Natural Language Inference}
	\acro{OSM}{OpenStreetMap}
	\acro{OWL}{Web Ontology Language}
	\acro{OOV}{out-of-vocabulary}
	\acro{PFM}{Pseudo-F-Measures}
	\acro{PSO}{Particle-Swarm Optimization}
	\acro{PBSMT}{Phrase-Based Statistical Machine Translation}
	
	\acro{QA}{Question Answering}
	\acro{RL}{Reinforcement Learning}
	\acro{RDF}{Resource Description Framework}
	\acro{RNN}{Recurrent Neural Network}
	\acro{ReLU}{rectified linear unit}
	
	\acro{SKOS}{Simple Knowledge Organization System}
	\acro{SWT}{Semantic Web Technologies}
	\acro{SW}{Semantic Web}
	\acro{SPARQL}{SPARQL Protocol and RDF Query Language}
	\acro{SRL}{Statistical Relational Learning}
	\acro{SVM}{Support Vector Machines}
    \acro{SGD}{Stochastic Gradient Descent}

    \acro{TBMT} {Transfer-Based Machine Translation}
	\acro{UML}{Unified Modeling Language}
	\acro{USL}{Ukrainian Sign Language}
	\acro{URI}{Uniform Resource Identifier}
	\acro{WHO}{World Health Organization}
	\acro{WKT}{Well-Known Text}
	\acro{W3C}{World Wide Web Consortium}
	\acro{WSD}{Word Sense Disambiguation}
    \acro{XML}{Extensible Markup Language}
	\acro{YPLL}{Years of Potential Life Lost}
\end{acronym}  

\newcommand{\kg}{\ensuremath{\mathcal{G}}\xspace}

\newcommand{\triple}[3]{(\texttt{#1}, \texttt{#2}, \texttt{#3})}

\newcommand{\entities}{\ensuremath{\mathcal{E}}\xspace}
\newcommand{\relations}{\ensuremath{\mathcal{R}}\xspace}

\newcommand{\scoreFunc}{\phi}

\copyrightyear{2021}
\acmYear{2021}
\acmConference[WWW '21]{Proceedings of the 2021 World Wide Web Conference}{April 19--23, 2021}{Ljubljana, Slovenia}
\acmBooktitle{Proceedings of the 2021 World Wide Web Conference (WWW '21), April 19--23, 2021, Ljubljana, Slovenia}
\acmPrice{}

\begin{document}

\title[Out-of-Vocabulary Entities in Link Prediction]{Out-of-Vocabulary Entities in Benchmarks for Link Prediction}

\author{Caglar Demir}
\affiliation{
\institution{Data Science Research Group, Paderborn University}
}
\author{Axel-Cyrille Ngonga Ngomo}
\affiliation{
\institution{Data Science Research Group, Paderborn University}
}

\begin{abstract}
Knowledge graph embedding techniques are key to making knowledge graphs amenable to the plethora of machine learning approaches based on vector representations. Link prediction is often used as a proxy to evaluate the quality of these embeddings. Given that the creation of benchmarks for link prediction is a time-consuming endeavor, most work on the subject matter uses only a few benchmarks. As benchmarks are crucial for the fair comparison of algorithms, ensuring their quality is tantamount to providing a solid ground for developing better solutions to link prediction and ipso facto embedding knowledge graphs. First studies of benchmarks pointed to limitations pertaining to information leaking from the development to the test fragments of some benchmark datasets. We spotted a further common limitation of three of the benchmarks commonly used for evaluating link prediction approaches: out-of-vocabulary entities in the test and validation sets. We provide an implementation of an approach for spotting and removing such entities and provide corrected versions of the datasets WN18RR, FB15K-237, and YAGO3-10. Our experiments on the corrected versions of WN18RR, FB15K-237, and YAGO3-10 suggest that the measured performance of state-of-the-art approaches is altered significantly with p-values $<1\%, <1.4\%$, and $<1\%$, respectively. Overall, state-of-the-art approaches gain on average absolute $3.29 \pm 0.24\%$ in all metrics on WN18RR. This means that some of the conclusions achieved in previous works might need to be revisited. We provide an open-source implementation of our experiments and corrected datasets at \url{https://github.com/dice-group/OOV-In-Link-Prediction}.
\end{abstract}
\maketitle
\section{Introduction}
\label{sec: introduction}

\acp{KG} represent structured collections of facts modelled in typed relationships between entities~\cite{hogan2020knowledge}. These collections of facts have been used in a wide range of applications, including web search \cite{eder2012knowledge}, cancer research \cite{saleem2014big}, and even entertainment \cite{malyshev2018getting}. However, most \acp{KG} on the Web are far from being complete~\cite{nickel2015review}.
For instance, birthplaces for $71\%$ of people in Freebase and $66\%$ of people in DBpedia are not found in the respective \acp{KG}. Additionally, more than $58\%$ of scientists in DBpedia are not linked to the predicate that describes what they are known for~\cite{krompass2015type}. Link prediction on \acp{KG} refers to identifying such missing information~\cite{dettmers2018convolutional}.~\ac{KGE} approaches have been particularly successful at tackling the link prediction task~\cite{nickel2015review}. Hence, link prediction is often used as a proxy to measure the quality of \ac{KGE} approaches.

To quantify link prediction performances of 
~\ac{KGE} approaches, the \ac{KGE} research community often relies on the datasets WN18, WN18RR, FB15K, FB15K-237, and YAGO3-10. Tautanova and Chen~\cite{toutanova2015representing} note the test leakage problem of FB15K and WN18, i.e., a large number of triples in the test sets can be derived by inverting triples in the train set~\cite{dettmers2018convolutional}. Tautanova and Chen~\cite{toutanova2015representing} construct FB15K-237 via removing near-duplicate and inverse-duplicate relations from FB15K. Dettmers et al.~\cite{dettmers2018convolutional} investigate the severity of this problem and show that a simple rule-based model exploiting the test leakage problem of FB15K and WN18 achieves state-of-the-art link prediction performance on both datasets. Dettmers et al.~\cite{dettmers2018convolutional} create WN18RR that cannot be exploited using a single rule and caution against using FB15K and WN18. Two recent studies point out that the same benchmark datasets may suffer from another problem: the validation and test sets of WN18RR, FB15K-237, and YAGO3-10 contain entities that do not occur during training~\cite{demir2021convolutional,libkge}.

The goal of this paper is two-fold: first, we aim to provide  new, corrected versions of all three datasets. We dub these corrected versions WN18RR$^{\star}$, FB15K-237$^{\star}$ and YAGO3-10$^{\star}$. Second, we investigate the impact of out-of-vocabulary entities (entities that do not occur in the train set) on the link prediction performance of a selection of state-of-the-art \ac{KGE} techniques. Our experiments indicate that discrepancies in link prediction performances of approaches are statistically significant with p-values $<1\%, <1.4\%$, and $<1\%$ on WN18RR, FB15K-237, and YAGO3-10 (see~\Cref{sec:evaluation}). For instance, approaches achieve absolute gains of $3.29 \pm 0.24\%$ on average, in all metrics on  WN18RR$^{\star}$, while the link prediction performance of TransE on FB15K-237$^{\star}$ is increased by absolute $9\%$ in all metrics. During our experiments, we observed that out-of-vocabulary entities often occur with particular relations (e.g., \texttt{hypernym} on WN18RR). Hence, previously reported link prediction per relation results do not fully reflect actual performances of approaches. In turn, we observe that the impact of out-of-vocabulary entities is not as severe in ranking missing relations as in ranking missing head and tail entities.
\section{Related work}
\label{sec:related work}
A wide range of works have investigated \ac{KGE} to address various tasks such as type prediction, relation prediction, link prediction, question answering, item recommendation and knowledge graph completion~\cite{demir2019physical,demir2021shallow,nickel2011three,huang2019knowledge}. We refer to~\cite{nickel2015review,wang2017knowledge,cai2018comprehensive,ji2020survey,qinsurvey} for recent surveys and briefly overview selected \ac{KGE} techniques.

RESCAL~\cite{nickel2011three} is a bilinear model that computes a three-way factorization of a third-order adjacency tensor representing the input \ac{KG}. RESCAL captures various types of relations in the input KG but is limited in its scalability as it has quadratic complexity in the  factorization rank~\cite{trouillon2017knowledge}. DistMult~\cite{yang2015embedding} can be seen as an efficient extension of RESCAL with a diagonal matrix per relation to reduce the complexity of RESCAL~\cite{balavzevic2019tucker}. DistMult performs poorly on antisymmetric relations while performing well on symmetric relations~\cite{trouillon2017knowledge}. Note that through applying the reciprocal data augmentation technique, this incapability of DistMult is alleviated~\cite{ruffinelli2019you}. TuckER~\cite{balavzevic2019tucker} performs a Tucker decomposition on the binary tensor representing the input \ac{KG} which enabling multi-task learning through parameter sharing between different relations via the core tensor. ComplEx~\cite{trouillon2016complex} extends DistMult by learning representations in a complex vector space. ComplEx can infer both symmetric and antisymmetric relations via a Hermitian inner product of embeddings that involves the conjugate-transpose of one of the two input vectors. ComplEx yields state-of-the-art performance on the link prediction task while leveraging the linear space and time complexity of the dot products. 
Trouillon et al.~\cite{trouillon2017complex} showed that ComplEx is equivalent to HolE~\cite{nickel2015holographic}. 
ConvE~\cite{dettmers2018convolutional} is a convolutional neural model that relies on a 2D convolution operation to capture the interactions between entities and relations. Through interactions captured by 2D convolution, ConvE yields a state-of-the-art performance in link prediction. ConEx~\cite{demir2021convolutional} combines a 2D convolution followed by an affine transformation with a Hermitian inner product of complex-valued embeddings to generate scores of triples. Hence, ConEx can be considered a multiplicative inclusion of ConvE into ComplEx.
\section{Preliminaries}
\label{sec:preliminaries}
\paragraph{Knowledge Graphs.} Let \entities\ and \relations\ represent the set of entities and relations, respectively. Then, a \ac{KG} $\kg= \{\triple{h}{r}{t}  \in \entities \times \relations \times \entities\}$ can be formalised as a set of triples where each triple contains two entities $\texttt{h},\texttt{t} \in \entities$ and a relation $\texttt{r} \in \relations$. 
\paragraph{Link Prediction.} The link prediction task addresses the problem of predicting whether unseen triples (i.e., triples not found in \kg) are true~\citep{ji2020survey}. The task is often formalised by learning a parameterised scoring function $\scoreFunc_\Theta:\entities \times \relations \times \entities \mapsto \mathbb{R}$~\citep{nickel2015review,ji2020survey} ideally characterised by $\scoreFunc_\Theta \triple{h}{r}{t} > \scoreFunc_\Theta \triple{x}{y}{z}$ if $\triple{h}{r}{t}$ is true and $\triple{x}{y}{z}$ is not. 
An approach's link prediction performance is evaluated using its ranking of missing entities in unseen triples as a proxy
~\citep{nickel2011three,dettmers2018convolutional,nickel2015review,trouillon2016complex}. Consequently, benchmark datasets for link prediction consist of disjoint three sets of triples denoted by $\kg^{\text{Train}}$, $\kg^{\text{Val}}$, and $\kg^{\text{Test}}$. 
Vocabulary of entities for the training phase can be defined as $\entities^{\text{Train}} = \{ x | \triple{x}{r}{t} \vee \triple{h}{r}{x}\in \kg^{\text{Train}} \}$. Similarly, the vocabulary of relations can be defined as $\relations^{\text{Train}} = \{ r | \triple{h}{r}{t}\in \kg^{\text{Train}} \}$. Analogously, vocabulary of entities and relations for validation and test phases can be obtained.
\paragraph{Evaluation metrics.} Link prediction performances of approaches are often measured via the filtered mean reciprocal rank (MRR) and Hits@N metrics based on missing head and tail entity rankings~\citep{dettmers2018convolutional,nickel2015review,trouillon2016complex}. Formally, the filtered MRR is defined as 
\begin{equation}
    \text{MRR} = \frac{1}{2|\kg^{\text{Test}}|} \sum_{\triple{h}{r}{t} \in \kg^{\text{Test}}} \Big( \frac{1}{\text{rank}(t |h,r)} + \frac{1}{\text{rank}(h |r,t)} \Big),
    \label{eq:mrr_for_entities}
\end{equation}
where $\text{rank}(t |h,r)$ (equivalently $\text{rank}(h |r,t)$) is computed in three consecutive steps. (a) Scores are computed--$\forall x \in \entities : \scoreFunc_\Theta \triple{h}{r}{x}$. (b) Scores of triples that occurred in train, validation, and test sets are filtered except for the input test triple \triple{h}{r}{t}. (c) Entities are ranked in descending order of corresponding scores. Given $\triple{h}{r}{t} \in \kg^{\text{Test}}$, $\text{rank}(t |h,r)$ denotes the rank of missing $\texttt{t}$ in the ordered entities. To further investigate link prediction performances of approaches, link prediction per relation performances are often measured as 
\begin{equation} 
    \text{MRR}_{r_i} = \frac{1}{2|\kg^{\text{Test}}|} \sum_{ (\texttt{h},\texttt{r}_i,\texttt{t}) \in \kg^{\text{Test}}} \Big( \frac{1}{\text{rank}(t | h, r_i )} + \frac{1}{\text{rank}(h | r_i ,t)} \Big).
    \label{eq:mrr_per_relation}
\end{equation}
~\Cref{eq:mrr_per_relation} categorises \ac{MRR} scores per relation~\cite{sun2019rotate,zhang2019quaternion,balazevic2019multi}. Similarly, the filtered Hits@N is defined as 
\begin{equation}
    \small
    \frac{1}{2|\kg^{\text{Test}}|} \sum_{\triple{h}{r}{t} \in \kg^{\text{Test}}} \Big( \mathcal{I}(\text{rank}(t |h,r) \leq N) + \mathcal{I}(\text{rank}(h |r,t) \leq N)\Big),
    \label{eq:hits_at_n}
\end{equation}
where the binary function $\mathcal{I}$ returns 1 if the condition is true, otherwise 0. 
Link prediction performances can also be measured in terms of missing relation prediction~\cite{wang2017knowledge}.
To this end,~\Cref{eq:mrr_for_entities} is altered as
\begin{equation}
    \text{MRR} = \frac{1}{|\kg^{\text{Test}}|} \sum_{\triple{h}{r}{t} \in \kg^{\text{Test}}} \Big( \frac{1}{\text{rank}(r |h,t)} \Big).
    \label{eq:mrr_for_relations}
\end{equation}
Analogous to~\Cref{eq:mrr_for_relations},~\Cref{eq:hits_at_n} is modified to quantify relation prediction performances~\cite{demir2021shallow}.
\section{Problem}
\paragraph{Learning a scoring function.}
To tackle the link prediction problem, the common practice is to learn a parameterized scoring function $\scoreFunc_\Theta : \entities \times \relations \times \entities \mapsto \mathbb{R}$, where $\Theta$ corresponds to parameters of the scoring function. $\Theta$ involves embeddings of entities and relations and may involve additional parameters depending on the architecture of the scoring function, e.g., kernels in convolution operation, affine transformations,  linear transformations or even  tensors~\cite{dettmers2018convolutional,demir2021convolutional,balavzevic2019hypernetwork,demir2021shallow,nguyen2018novel,balavzevic2019tucker,nickel2011three}. The standard workflow of learning $\scoreFunc_\Theta$ to tackle the link prediction problem consists of three parts: 
\begin{enumerate}
    \item learning $\Theta$ by minimizing a set loss function (e.g., a margin or entropy-based loss functions) on $\kg^{\text{Train}}$,
    \item determining hyperparameters of $\scoreFunc_\Theta$ on $\kg^{\text{Val}}$, and
    \item measuring the link prediction performance of $\scoreFunc_\Theta$ on $\kg^{\text{Test}}$.
\end{enumerate}
This workflow entails that $\entities^{\text{Train}}$ and $\relations^{\text{Train}}$ are \emph{known a priori} since $\Theta$ must involve embeddings of all entities and relations seen during the training phase. Hence, $\Theta$ can be learned as a byproduct of minimizing the set loss function through an optimizer (e.g. ADAM~\cite{kingma2014adam}, RMSprop~\cite{Tieleman2012}). In this workflow, the following two conditions must hold:
\begin{enumerate}
    \item $\entities^{\text{Val}} \subseteq \entities^{\text{Train}} \wedge \entities^{\text{Test}} \subseteq \entities^{\text{Train}}$ and
    \item $\relations^{\text{Val}} \subseteq \relations^{\text{Train}} \wedge \relations^{\text{Test}} \subseteq \relations^{\text{Train}}$. 
\end{enumerate}
An OOV-entity is an entity $e \in (\entities^{\text{Val}} \backslash \entities^{\text{Train}} \cup \entities^{\text{Test}} \backslash \entities^{\text{Train}})$. OOV relations are defined analogously. If these two conditions do not hold, then approaches may not be able to compute the score of triples \triple{h}{r}{t} $\in \kg^{\text{Test}}$, as an embedding of an OOV-entity or OOV-relation is not learned. Surprisingly, the former condition does not hold on WN18RR, FB15K-237 and YAGO3-10. To the best of our knowledge, this issue is only recently mentioned~\cite{demir2021convolutional,libkge}. However, the impact of OOV-entities has not been investigated. In this work, we are interested in the impact of OOV-entities in link prediction and relation prediction.~\Cref{table:all_dataset_info} provides an overview of benchmark datasets. ~\Cref{table:ovv_in_datasets} reports statistics of out-of-vocabulary entities on WN18RR, FB15K-237 and YAGO3-10 benchmark datasets. Results indicate that (1) \textbf{$7\%$ validation and test splits of WN18RR dataset contain OOV-entities}, and (2) the number of OOV-entities increases as the out-degree of node/entity in $\kg^{\text{Train}}$ increases.

\begin{table}[tb]
    \caption{Overview of WN18RR, FB15K-237 and YAGO3-10 benchmark datasets. 
    $|\kg|$, $|\entities|$, and $|\relations|$ denote number of triples, entities, and relation.
    Indegr. and Outdegr. (M$\pm$SD) $\kg$ stand for mean and standard deviation of node indegrees and node outdegrees, respectively.}
    \centering
    \begin{tabular}{@{}l c c c}
    \toprule
                                            & WN18RR             & FB15K-237            & YAGO3-10 \\   
    \toprule
    $|\kg^{\text{Train}}|$                  & 86,835             & 272,115              & 1,079,040         \\
    $|\entities^{\text{Train}}|$            & 40,559             & 14,505               & 123,143       \\
    $|\relations^{\text{Train}}|$           & 11                 & 237                  & 37           \\
    Indegr. (M$\pm$SD) $\kg^{\text{Train}}$ & 2.72$\pm$7.74      & 20.34$\pm$98.54      & 22.51$\pm$293.96  \\
    Outdegr. (M$\pm$SD) $\kg^{\text{Train}}$& 2.19$\pm$3.56      & 19.746$\pm$30.10     & 9.56$\pm$8.67\\
    \midrule
    $|\kg^{\text{Val}}|$                    & 30,34              & 17,535               & 5000 \\
    $|\entities^{\text{Val}}|$              & 5,173              & 9,809                & 7948                    \\
    $|\relations^{\text{Val}}|$             & 11                 & 223                  & 33  \\
    Indegr. (M$\pm$SD) $\kg^{\text{Val}}$   & 1.18$\pm$0.87      & 3.02$\pm$11.76       & 1.59$\pm$5.25     \\
    Outdegr. (M$\pm$SD) $\kg^{\text{Val}}$  & 1.06$\pm$0.41      & 2.29$\pm$2.75        & 1.03$\pm$0.19     \\
    \midrule
    $|\kg^{\text{Test}}|$                   & 3,134              & 20,466               & 5000              \\
    $|\entities^{\text{Test}}|$             & 5,323              & 10,348               & 7937              \\
    $|\relations^{\text{Test}}|$            & 11                 & 224                  & 34                \\
    Indegr. (M$\pm$SD) $\kg^{\text{Test}}$  & 1.20$\pm$0.95      & 3.21$\pm$12.91       & 1.57$\pm$5.06     \\
    Outdegr. (M$\pm$SD) $\kg^{\text{Test}}$ & 1.06$\pm$0.44      & 2.50$\pm$3.20        & 1.04$\pm$0.21     \\
    \bottomrule
    \end{tabular}
    \label{table:all_dataset_info}
\end{table}

\begin{table}
    \caption{Overview of out-of-vocabulary entities in WN18RR, FB15K-237 and YAGO3-10 benchmark datasets.
    OOV in $\kg^{\text{.}}$ denotes number of triples containing at least an out-of-vocabulary entity and its percentage in the respective set.}
    \centering
    \begin{tabular}{l c c c}
    \toprule
                                                        & WN18RR        & FB15K-237    & YAGO3-10 \\
    \midrule
    $|\entities^\text{Test} - \entities^\text{Train}|$  & 209           & 29           & 18      \\
    OOV in $\kg^{\text{Test}}$                         & 210 ($6.70\%$)  & 28 ($0.14\%$)  & 18 ($0.36\%$)      \\
    \midrule
    $|\entities^\text{Val} - \entities^\text{Train}|$   & 198           & 8            & 22      \\
    OOV in $\kg^{\text{Val}}$                          & 210 ($6.92\%$)  & 9 ($0.05\%$)   & 22 ($0.44\%$)      \\
    \bottomrule
    \end{tabular}
    \label{table:ovv_in_datasets}
\end{table}

The publicly available implementations of state-of-the-art \ac{KGE} approaches often mitigate the issue of OOV-entities by constructing $\entities$ and $\relations$ using the train, validation and test sets~\cite{dettmers2018convolutional,trouillon2016complex,balavzevic2019tucker,balavzevic2019hypernetwork,balazevic2019multi,demir2021convolutional}. This entails that link prediction performances of \ac{KGE} approaches are quantified using embeddings of OOV-entities that are \textbf{not learned but solely initialized}. 
OOV-entities may lead the following undesired situations:
\begin{enumerate}
    \item previously reported link prediction performances do not reflect actual performances,
    \item link prediction per relation results biased towards relations do not occur with OOV entities, and
    \item the impact of initialization of embeddings is unintentionally magnified.
\end{enumerate}
In our experiments, we are interested in quantifying the severity of these situations.

\section{Experimental Setup}
\label{sec:experiments}
\paragraph{Baselines.}
In our experiments, we relied on pretrained RESCAL, TransE, DistMult, ComplEx, ConvE, ConEx, and TuckER approaches provided in~\cite{ruffinelli2019you,libkge,demir2021convolutional}.
This decision stems from the fact that Ruffinelli et al.~\cite{ruffinelli2019you} conducted an extensive analysis on the impact of hyperparameter optimization and training strategies in link prediction performances. Their findings indicate that the relative performance differences between various approaches often shrink and sometimes even reverse when compared to prior results, provided that approaches are optimized properly. Hence, we alleviated the impact of different training strategies in our experiments by relying on the results of~\cite{ruffinelli2019you}. 

%
\paragraph{Datasets.} Bordes et al.~\cite{bordes2013translating} construct WN18 and FB15K datasets to quantify the link prediction performance of \ac{KGE} approaches. WN18 is a subset of WordNet~\cite{miller1995wordnet} that contains lexical relations, while FB15K is a subset of Freebase~\cite{bollacker2008freebase}. Toutanova and Chen~\cite{toutanova2015observed} highlight the test leakage problem of benchmark datasets and construct the FB15K-237 dataset by removing triples containing near-duplicate and inverse-duplicate relations from the FB15K dataset. The FB15K-237 dataset is constructed by limiting the set of relations in FB15K to the most frequent 401 relations. Next, near-duplicate and inverse relations are detected and removed from the train set of FB15K. This process detects many inverse relationships in relations, e.g., \texttt{award\_award\_nominee} is inverse of \texttt{award\_nominee/award}. Similarly, Dettmers et al.~\cite{dettmers2018convolutional} created WN18RR, on which link prediction cannot easily be exploited by using a single rule.
\paragraph{Evaluation metrics}
We employ the standard metrics \textit{filtered} \ac{MRR} and hits at N (H@N) for link prediction and relation prediction~\cite{dettmers2018convolutional,trouillon2016complex,demir2021shallow}.
\section{Evaluation}
\label{sec:evaluation}
\begin{table*}[!t]
    \caption{Link prediction results on WN18RR and WN18RR$\star$.}
	\centering
	\small
    \begin{tabular}{lcccccccc}
    \toprule 
    &\multicolumn{4}{c}{\bf WN18RR}&\multicolumn{4}{c}{\bf WN18RR$\star$}\\ 
    \cmidrule(lr){2-5}\cmidrule(lr){6-9}

             & MRR  & Hits@1 & Hits@3 & Hits@10  & MRR  & Hits@1 & Hits@3 & Hits@10 \\
    \midrule
	RESCAL   & .467 & .439   & .480   & .517     & .499 & .470   & .513   & .552    \\
 	TransE   & .175 & .044   & .227   & .484     & .187 & .047   & .243   & .517    \\
	DistMult & .452 & .413   & .466   & .530     & .483 & .442   & .499   & .566    \\
	ComplEx  & .475 & .438   & .490   & .547     & .509 & .470   & .525   & .587    \\
	ConvE    & .442 & .411   & .451   & .504     & .472 & .440   & .482   & .538    \\
	ConEx    & .481 & .448   & .493   & .550     & .512 & .477   & .525   & .584    \\
	TuckER   & .466 & .441   & .476   & .515     & .488 & .471   & .509   & .550    \\
    \bottomrule
    \end{tabular}
     \label{table:wn18rr_and_wn18rr_star}
 \end{table*}

 \begin{table*}[!t]
    \caption{Link prediction results on FB15K-237 and FB15K-237$\star$.}
	\centering
	\small
    \begin{tabular}{lcccccccc}
    \toprule 
    &\multicolumn{4}{c}{\bf FB15K-237}&\multicolumn{4}{c}{\bf FB15K-237$\star$}      \\ 
    \cmidrule(lr){2-5}\cmidrule(lr){6-9}
             & MRR   & Hits@1 & Hits@3 & Hits@10 & MRR  & Hits@1 & Hits@3 & Hits@10   \\
    \midrule
	RESCAL   & .351  & .260   & .387   & .531   & .354  & .262   & .391   & .537       \\
	TransE   & .150  & .090   & .148   & .284   & .246  & .175   & .263   & .390       \\
	DistMult & .343  & .249   & .378   & .531   & .343  & .249   & .379   & .532       \\
	ComplEx  & .348  & .253   & .384   & .536   & .348  & .253   & .384   & .536       \\
	ConvE    & .329  & .244   & .359   & .501   & .334  & .246   & .364   & .501       \\
	ConEx    & .366  & .271   & .403   & .555   & .366  & .271   & .403   & .555        \\
	TuckER   & .363  & .268   & .400   & .553   & .363  & .268   & .400   & .553        \\
    \bottomrule
    \end{tabular}
     \label{table:fb15k3_27_and_fb15k3_27_star}
 \end{table*}
 
 \begin{table*}[!t]
    \caption{Link prediction results on YAGO3-10 and YAGO3-10$\star$.}
	\centering
	\small
    \begin{tabular}{lcccccccc}
    \toprule 
    &\multicolumn{4}{c}{\bf YAGO3-10}&\multicolumn{4}{c}{\bf YAGO3-10$\star$}\\ 

             & MRR   & Hits@1 & Hits@3 & Hits@10 & MRR  & Hits@1 & Hits@3 & Hits@10   \\
    \midrule
	DistMult & .543  & .466   & .590   & .683   & .545  & .467   & .592   & .686        \\
	ComplEx  & .547  & .468   & .594   & .690   & .549  & .470   & .596   & .692        \\
	ConEx    & .553  & .474   & .601   & .696   & .555  & .476   & .603   & .698        \\
	TuckER   & .427  & .331   & .476   & .609   & .429  & .332   & .477   & .611        \\
    \bottomrule
    \end{tabular}
     \label{table:yago3_10_and_yago3_10_star}
 \end{table*}

\begin{table*}[!t]
    \caption{Relation prediction results on WN18RR and WN18RR$\star$.}
	\centering
	\small
    \begin{tabular}{lcccccccc}
    \toprule 
    &\multicolumn{4}{c}{\bf WN18RR}&\multicolumn{4}{c}{\bf WN18RR$\star$}\\ 
    \cmidrule(lr){2-5}\cmidrule(lr){6-9}
                                      & MRR  & Hits@1 & Hits@3 & Hits@10  & MRR  & Hits@1 & Hits@3 & Hits@10 \\
    \midrule
    RESCAL                            & .578 &  .335  &  .784  &  .892    & .587 &  .343   &  .797 &  .891   \\
    TransE                            & .547 &  .360  &  .682  &  .865    & .525 &  .328   &  .669 &  .859   \\
    DistMult                          & .671 &  .539  &  .759  &  .869    & .687 &  .553   &  .780 &  .889   \\
	ComplEx                           & .785 &  .705  &  .822  &  .989    & .820 &  .749   &  .857 &  .994   \\
    ConvE                             & .353 &  .143  &  .405  &  .857    & .358 &  .148   &  .414 &  .853   \\
    \bottomrule
    \end{tabular}
    \label{table:relation_prediction_wn}
 \end{table*}
 
 \begin{table*}[!t]
    \caption{Relation prediction results on FB15K-237 and FB15K-237$\star$.}
	\centering
	\small
    \begin{tabular}{lcccccccc}
    \toprule 
    &\multicolumn{4}{c}{\bf FB15K-237}&\multicolumn{4}{c}{\bf FB15K-237$\star$}\\ 
    \cmidrule(lr){2-5}\cmidrule(lr){6-9}
                                      & MRR  & Hits@1 & Hits@3 & Hits@10 & MRR   & Hits@1 & Hits@3 & Hits@10 \\
    \midrule
    RESCAL                            & .192 &  .021  &  .140  &  .823   &  .192 &  .021  &  .141  &  .824   \\
    TransE                            & .672 &  .589  &  .733  &  .800   &  .673 &  .589  &  .734  &  .801    \\
    DistMult                          & .568 &  .425  &  .660  &  .823   &  .569 &  .425  &  .661  &  .824   \\
	ComplEx                           & .632 &  .506  &  .717  &  .855   &  .633 &  .507  &  .718  &  .856   \\
    ConvE                             & .667 &  .562  &  .732  &  .874   &  .667 &  .562  &  .732  &  .874   \\
    \bottomrule
    \end{tabular}
    \label{table:relation_prediction_fb}
 \end{table*}
 
    
%
\Cref{table:wn18rr_and_wn18rr_star,table:fb15k3_27_and_fb15k3_27_star,table:yago3_10_and_yago3_10_star} report link prediction results on WN18RR, FB15K-237, YAGO3-10 and their respective corrected versions. Our results on WN18RR$^{\star}$ indicate that state-of-the-art approaches achieve on average absolute $3.29 \pm 0.24\%$ gains in all metrics.
Such a distinct difference is not observed on FB15K-237$^{\star}$ and YAGO3-10$^{\star}$.

These results also indicate that shallow models (e.g., DistMult and ComplEx) perform particularly well on FB15K-237, YAGO3-10, and their corrected versions. These results hence do not corroborate the claim that shallow models do not perform well on knowledge graphs having high node in-degree~\cite{dettmers2018convolutional}. In turn, we conjecture that this claim might have to be reconsidered. Although on average in-degree of nodes in FB15K-237 is circa 7.5 times larger than WN18RR, ComplEx and DistMult outperform ConvE in all metrics. In FB15K-237$^{\star}$, the link prediction performances of RESCAL, TransE, and ConvE are increased in all metrics, whereas link prediction performances of DistMult and ComplEx did not change.
\paragraph{Link Prediction per Relation.} During our evaluation, we were interested in quantifying the impact of OVV-entities in the link prediction per relation task.
~\Cref{table:lp_per_relation_wn,table:lp_per_relation_fb,table:lp_per_relation_yago} report link prediction per relation results on all datasets. Results indicate that performances of approaches improve particularly well on \texttt{hypernym}, \texttt{instance\_hypernym}, \texttt{member\_meronym}, \texttt{synset\_domain\_topic\_of}, \texttt{has\_part} and \text{member\_of\_domain\_usage} on WN18RR$^\star$, whereas such distinct improvements are not observed on the remaining relations. 
\begin{table*}[!t]
\caption{MRR link prediction per relation results on WN18RR and WN18RR*.}
\centering
\small
\begin{tabular}{@{}lcccccccccc@{}}
\toprule
                                       & RESCAL    & TransE    & DistMult & ComplEx    & ConvE   & ConEx    & TuckER \\
\midrule
\bf WN18RR \\ \midrule
hypernym                               & .149      & .071      & .117     & .157       & .090    & .150     & .121      \\
instance\_hypernym                     & .349      & .246      & .377     & .385       & .352    & .392     & .375      \\
member\_meronym                        & .225      & .168      & .148     & .232       & .198    & .171     & .181      \\
synset\_domain\_topic\_of              & .325      & .301      & .348     & .335       & .311    & .373     & .344      \\
has\_part                              & .164      & .095      & .155     & .212       & .146    & .192     & .171      \\
member\_of\_domain\_usage              & .295      & .328      & .308     & .215       & .299    & .319     & .212      \\
member\_of\_domain\_region             & .193      & .228      & .365     & .120       & .367    & .354     & .284      \\
derivationally\_related\_form          & .956      & .276      & .957     & .957       & .953    & .986     & .985      \\
also\_see                              & .595      & .209      & .613     & .616       & .656    & .647     & .658      \\
verb\_group                            & .974      & .337      & .975     & .974       & .974    & 1.00     & .987      \\
similar\_to                            & .585      & .285      & 1.00     & 1.00       & 1.00    & 1.00     & 1.00      \\
\midrule
\bf WN18RR$^{\star}$\\ \midrule
hypernym                               & .173      & .083      & .135     & .184       & .105    & .169     & .140      \\
instance\_hypernym                     & .375      & .259      & .407     & .419       & .376    & .423     & .408      \\
member\_meronym                        & .231      & .173      & .152     & .238       & .203    & .176     & .186      \\
synset\_domain\_topic\_of              & .333      & .312      & .355     & .347       & .318    & .376     & .350      \\
has\_part                              & .167      & .097      & .158     & .216       & .148    & .191     & .169      \\
member\_of\_domain\_usage              & .303      & .329      & .308     & .234       & .307    & .319     & .232      \\
member\_of\_domain\_region             & .193      & .229      & .365    & .120        & .367    & .354     & .284      \\
derivationally\_related\_form          & .956      & .276      & .957    & .957        & .953    & .986     & .985      \\
also\_see                              & .594      & .209      & .613    & .616        & .656    & .647     & .658      \\
verb\_group                            & .974      & .337      & .975    & .974        & .974    & 1.00     & .987      \\
similar\_to                            & .585      & .285      & 1.00    & 1.00        & 1.00    & 1.00     & 1.00      \\
\bottomrule
\end{tabular}
\label{table:lp_per_relation_wn}
\end{table*}

\begin{table*}[!t]
\caption{MRR link prediction per relation results based on some person related relations.}
\centering
\small
\begin{tabular}{@{}lcccccccccc@{}}
\toprule
                                       & RESCAL    & TransE   & DistMult & ComplEx    & ConvE  & ConEx      & TuckER \\
\midrule
\bf FB15K-237 \\ \midrule 
people/person/profession               & .371      &.134      & .388     & .402       & .386   & .374      & .375       \\
people/person/gender                   & .585      &.452      & .528     & .518       & .505   & .593      & .605       \\
people/person/nationality              & .461      &.370      & .444     & .439       & .442   & .468      & .469       \\
people/person/languages                & .386      &.322      & .430     & .429       & .420   & .441      & .431       \\
people/person/religion                 & .306      &.264      & .301     & .302       & .304   & .309      & .325       \\
\midrule
\bf FB15K-237$^{\star}$\\ \midrule
people/person/profession               & .377      &.340      & .388     & .402       & .386   & .374      & .375       \\
people/person/gender                   & .586      &.452      & .528     & .518       & .505   & .593      & .605       \\
people/person/nationality              & .466      &.399      & .444     & .439       & .442   & .468      & .469       \\
people/person/languages                & .422      &.401      & .430     & .429       & .421   & .441      & .431       \\
people/person/religion                 & .307      &.283      & .301      & .302      & .304   & .309      & .325       \\
\bottomrule
\end{tabular}
\label{table:lp_per_relation_fb}
\end{table*}

\begin{table*}[!t]
\caption{MRR link prediction per relation results based on some person and place related relations.}
\centering
\small
\begin{tabular}{@{}lcccccccc@{}}
\toprule
                                  & DistMult & ComplEx  & ConEx & TuckER  \\
\midrule
\bf YAGO3-10 \\ \midrule
isLocatedIn                      & .293          &.289         & .354       & .368       \\
happenedIn                       & .505         & .499         & .473     & .434       \\
directed                         & .509         & .504         & .527       & .480       \\
hasWonPrize                      & .245         & .237         & .281       & .269       \\
isMarriedTo                      & .985        & .985         & .985     & .936       \\
hasCapital                       & .483        & .481         & .564      & .442       \\
hasNeighbor                      & 1.00        & 1.00         & 1.00      & .406       \\
\midrule
\bf YAGO3-10$^{\star}$\\ \midrule
isLocatedIn                      & .302        & .297         & .365     & .379       \\
happenedIn                       & .505        & .499         & .473     & .434       \\
directed                         & .509        & .504         & .528     & .480       \\
hasWonPrize                      & .247        & .237         & .282     & .271       \\
isMarriedTo                      & .985        & .985         & .985     & .936       \\
hasCapital                       & .571        & .568         & .667     & .523       \\
hasNeighbor                      & 1.00        & 1.00         & 1.00     & .406       \\
\bottomrule
\end{tabular}
\label{table:lp_per_relation_yago}
\end{table*}
\paragraph{Statistical Hypothesis Testing.}
We carried out a Wilcoxon signed-rank test to check whether the impact of OOV entities in link prediction performances is significant. Our null hypothesis was that the link prediction performances of state-of-the-art approaches on the datasets with and without OOV entities and relations come from the same distribution. The alternative hypothesis was that these results come from different distributions, i.e., removing triples containing OOV entities from the test set has a significant impact on link prediction performances. To perform the Wilcoxon signed-rank test (two-sided), we used the
\ac{MRR}, Hits@1, Hits@3, and Hits@10 performances on a benchmark dataset and its corrected version (e.g. WN18RR-WN18RR$\star$, FB15K-237-FB15K-237$\star$, and YAGO3-10-YAGO3-10$\star$). We were able to reject the null hypothesis with p-values 
$<1\%, <1.4\%$, and $<1\%$, 
on all three tests.
\section{Conclusion}
\label{sec:conclusion}
In this work, we investigated the impact of out-of-vocabulary entities in the link prediction and relation prediction problem. Our analysis shows that $6.70\%$ and $6.92\%$ triples in the test and validation splits of the WN18RR benchmark dataset do not serve for benchmarking link prediction performances of approaches. Our experiments also showed that state-of-the-art approaches gain on average absolute $3.29 \pm 0.24\%$ in all metrics on WN18RR. Findings of our statistically hypothesis test indicates that link prediction performances of state-of-the-art approaches are significantly altered. In turn, the impact of out-of-vocabulary entities in ranking missing relations is not as distinct as in ranking missing entities. 

We provide provide an open-source implementation of our experiments along with corrected datasets in the project page.

{\raggedright
  \bibliographystyle{ACM-Reference-Format}
  \bibliography{ref}}
\end{document}